\documentclass[letterpaper, 10 pt, conference]{ieeeconf}
\IEEEoverridecommandlockouts
\usepackage{cite}
\usepackage{amsmath,amssymb,amsfonts}
\usepackage{algorithmic}
\usepackage{graphicx} 
\usepackage{subfigure}
\usepackage{wrapfig}
\usepackage{textcomp}
\usepackage{booktabs}
\usepackage{tabulary}
\usepackage{multirow}
\usepackage{dirtytalk}
\usepackage{url}
\usepackage{afterpage}
\usepackage{caption}

\begin{document}

\title{\LARGE \bf Robot-Assisted Nuclear Disaster Response: Report and Insights from a Field Exercise}

\author{Manolis Chiou\textsuperscript{2}, Georgios-Theofanis Epsimos\textsuperscript{1}, Grigoris Nikolaou\textsuperscript{1}, Pantelis Pappas\textsuperscript{1}, Giannis Petousakis\textsuperscript{2},\\ Stefan Mühl\textsuperscript{3}, and Rustam Stolkin\textsuperscript{2}

\thanks{This work was supported by the UKRI-EPSRC grant EP/R02572X/1 (UK National Centre for Nuclear Robotics).}%
\thanks{\textsuperscript{1}University of West Attica, Greece,
        {\tt\small nikolaou@uniwa.gr}}%
\thanks{\textsuperscript{2}Extreme Robotics Lab, University of Birmingham, UK,
        {\tt\small m.chiou@bham.ac.uk}}%
\thanks{\textsuperscript{3}Kerntechnische Hilfdienst GmbH (KHG), Germany}
}

\maketitle
\thispagestyle{empty}
\pagestyle{empty}


\begin{abstract}

This paper reports on insights by robotics researchers that participated in a 5-day robot-assisted nuclear disaster response field exercise conducted by Kerntechnische Hilfdienst GmbH (KHG) in Karlsruhe, Germany. The German nuclear industry established KHG to provide a robot-assisted emergency response capability for nuclear accidents. We present a systematic description of the equipment used; the robot operators' training program; the field exercise and robot tasks; and the protocols followed during the exercise. Additionally, we provide insights and suggestions for advancing disaster response robotics based on these observations. Specifically, the main degradation in performance comes from the cognitive and attentional demands on the operator. Furthermore, robotic platforms and modules should aim to be robust and reliable in addition to their ease of use. Last, as emergency response stakeholders are often skeptical about using autonomous systems, we suggest adopting a variable autonomy paradigm to integrate autonomous robotic capabilities with the human-in-the-loop gradually. This middle ground between teleoperation and autonomy can increase end-user acceptance while directly alleviating some of the operator's robot control burden and maintaining the resilience of the human-in-the-loop. 

\end{abstract}


\section{Introduction}


Disasters and extreme incidents, either man-made (e.g., industrial and nuclear accidents) or natural (e.g., floods, earthquakes, or wildfires), significantly impact lives, the economy, and the environment. Robotic systems deployed in areas of interest before, during, or after a disaster can aid First Responders (FRs), the affected population, and other stakeholders by mitigating many risks and costs. Robots can be used as active mobile sensing platforms or as embodied AI agents acting in the environment. FRs can primarily benefit from the ability to efficiently sense and act at a safe distance from the disaster site \cite{Murphy2014_disasterBook, Murphy2008_shared_role_model}. Given their potential benefits, robotic systems are increasingly deployed in disasters \cite{Murphy2014_disasterBook, Murphy2019_off_normal, Kruijff2012_Miradona_earthquake, Kruijff-Korbayova2016_Amatrice_earthquake, Surmann2021_fireBerlin}, aspiring to become a common asset of emergency response capabilities \cite{Kruijff-Korbayova2021_DRZ_center}. 

For this to happen, a two-way knowledge transfer from the robotic research community to stakeholders and from stakeholders and end-users to the researchers is necessary. The research community and stakeholders/end-users have different knowledge and skills, often complementary and synergistic. Stakeholders and end-users have extensive expert knowledge, field experience, and tested methods and procedures, often built up over many years. Robotic researchers should aim to learn from this experience and understand the constraints and requirements of the end-users before developing new technologies. In this way, the robotics research community can deliver relevant and valuable technological advances for stakeholders and be systematically evaluated in safe but realistic ways, e.g., deployed in field exercises. 

Towards this end, in this paper, we are contributing to the existing body of work that reports insights from field deployments of robotic systems in disaster response and realistic field exercises. We present a systematic description of a 5-day nuclear disaster response field exercise organized by the German nuclear disaster response organization Kerntechnische Hilfdienst GmbH (KHG). We participated in the exercise as robotics researchers and robot operator trainees. Additionally, we contribute by providing insights on the difficulties encountered by the robot operators; the current needs, culture, and procedures of the stakeholders; and suggestions on advancing disaster response robotic systems. We propose the careful and targeted integration of more autonomous capabilities into the robotic systems via a Variable autonomy (VA) paradigm while retaining the human-in-the-loop. We aim for these suggestions and insights to be valuable and relevant to the robotics community and the end-users.

\section{Related Work}

This section aims to provide a brief overview of robot-assisted nuclear disaster response, as a complete survey is outside the scope of this paper. The reader is referred to the comprehensive literature survey of Marques et al. \cite{Marques2021} for the state-of-the-art in mobile radiation detection systems (e.g., sensors and robotic platforms). The reader can also consult the following papers on robotic disaster response surveys from the perspectives of rescue robotics \cite{delmerico2019current}, drones \cite{daud2022applications}, and crisis management  \cite{Wilk-Jakubowski2022_crisis_management}. 

This paper aims to address the lack of robot-assisted nuclear response reporting created by the rarity of those incidents. Related work offering a detailed report on robot-assisted nuclear disaster response is mainly limited to the robot deployment in the Fukushima Daiichi nuclear disaster \cite{Kawatsuma2012_Fuku_robots_lessons_brief, Nagatani2013_Fuku_QuinceRobot, fuku_blog}. The main tasks of the robotic missions were exploration, inspection, measuring radiation levels, and taking samples of the environment. Additionally, Sato et al. \cite{doi:10.1080/00223131.2019.1581111}, conducted radiation imaging at the Fukushima Daiichi reactor buildings using a crawler robot. Similar work in robotic field deployment at a nuclear disaster site is the work of Connor et al. \cite{Connor2020} in which they used an unmanned aerial vehicle (UAV) to map the radiation dosage in the Chernobyl Exclusion Zone.

In general, research on mobile robots in the nuclear field is focused on the "prevention \& preparation" and "recovery" phases of the disaster cycle. Applications such as monitoring \cite{bird2018robot}, inspection \cite{friedrich2009miniature}, data collection, and sampling \cite{ducros2017rica} are the most prevalent use of mobile platforms. There is a trend to delegate mapping and radiation detection to drones \cite{Connor2020}.


When exposed to high radiation levels, robots tend to break down, causing issues in the exploration of the area and possibly blocking pathways, impeding future exploration. Groves et al. \cite{Groves2021} propose a radiation detection module and navigation package that uses the collected radiation information to augment the navigation protocol of the robot by updating the cost maps and reducing breakdown risk.
In the field of sensing, Vetter et al. \cite{Vetter2019}, propose a module that can blend data from cameras and radiation sensors to create a 3D representation of the radiation sources in an area. It could allow users to overlay radiation readings on a particular region. 
Lastly, since actual nuclear facilities are often off-limits due to various factors (e.g., restrictions due to security and radiation exposure), the work of Wright et al. \cite{Wright2021} can be valuable for research. They simulate Ionizing Radiation sensors and sources in the gazebo simulator, allowing further testing in simulation to develop applications suited for field deployment.

\section{The KHG organization}

 Kerntechnische Hilfdienst GmbH (KHG), the German nuclear disaster response organization, was founded in 1977 by the companies operating nuclear power plants in Germany, together with the fuel cycle industry and major research centers. KHG's mission is part of the emergency precautions and planning taken to stabilize a nuclear plant following an accident or breakdown, analyze the cause, and eliminate the effects. KHG is responsible for low-risk routine tasks such as simple measurements of radiation levels and minor or major nuclear disaster incidents and accidents. Their main areas of responsibility are: 

\begin{itemize}
    \item Radiation measurements inside and outside of nuclear facilities.
    \item Recovering of radioactive materials.
    \item Inspection and work at locations with maximum dose. allowance rates using remote-controlled manipulator vehicles
    \item Decontamination of personnel, equipment, and enclosed areas.
    \item Filtering waste air with mobile equipment.
    \item Collection of low-level radioactive wastewater.
\end{itemize}

Most of these procedures are either fully or partially robot-assisted.

\subsection{Organization's General Response Protocol}

KHG provides a response team that is 24/7 on-call upon their shareholders' request (e.g., the German nuclear industry). In addition to the response team, approximately 140 skilled personnel from service companies are trained by KHG for a response. KHG's equipment is on wheels and portable, so if an alarm goes off, they need approximately a couple of hours to gather people, pack the required equipment, and depart. The organization's general response protocol: 

\begin{enumerate}
    \item Receive a call at the operational call center.
    \item Locate the approximate area of the event.
    \item All relevant team members and personnel are gathered.
    \item An initial risk assessment of the incident is being made.
    \item Gathering of the suitable equipment.
    \item Departure towards the incident.
\end{enumerate}

As a general rule, the head of the operation is a supervisor that KHG appoints. The head of the operation is responsible for the safety of the crew and equipment as well as for the operation plan. If needed, they can intervene during the robot-assisted tasks. Details of the exercise-related protocol can be found in consequent sections.

\begin{figure*}
    \begin{minipage}[l]{1.0\columnwidth}
        \centering
        \includegraphics[width=0.85\columnwidth]{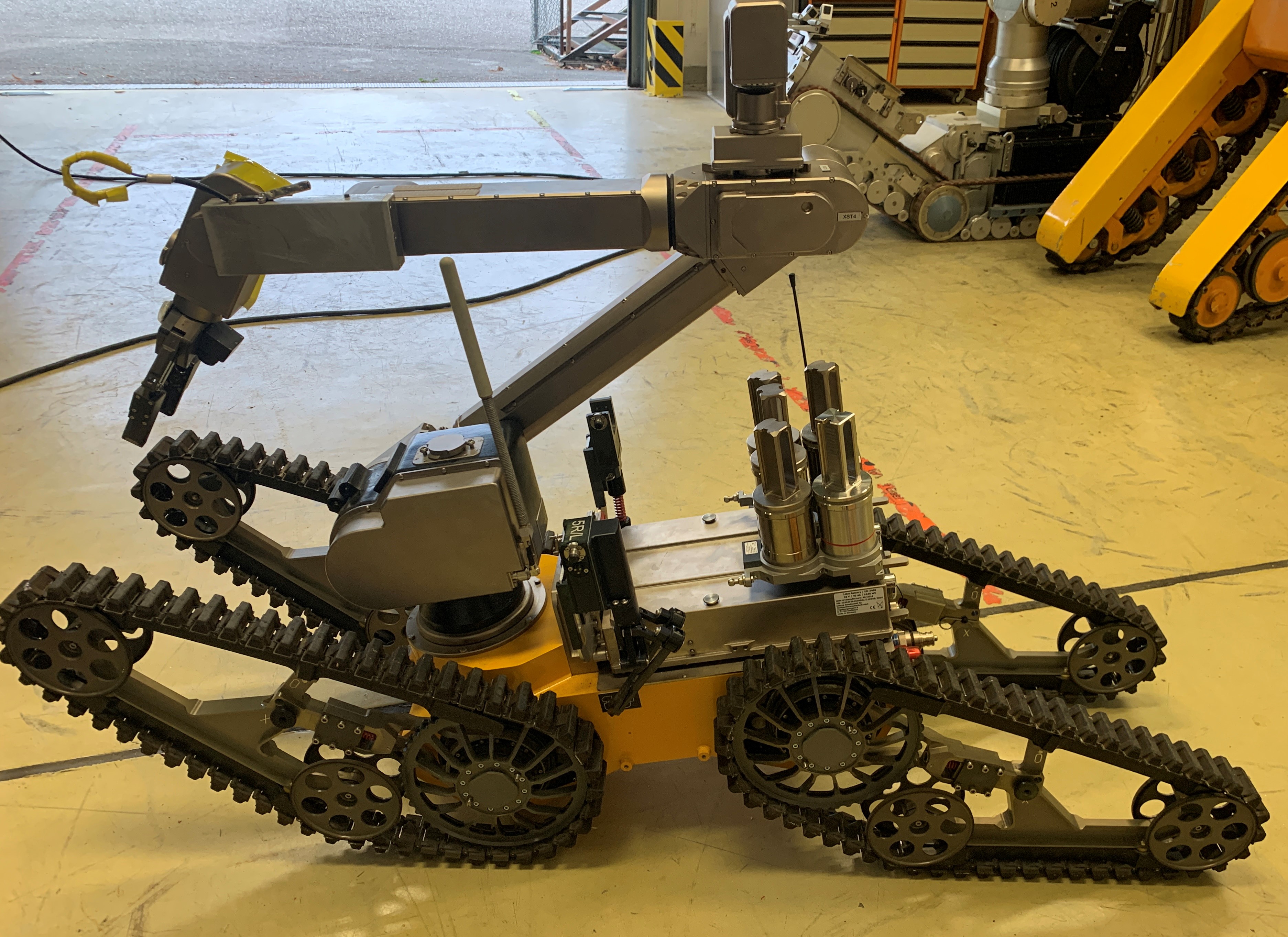}
        \caption{The Telemax mobile manipulator robot with the on-board sampling probes in the back.}\label{fig:Telemax}
    \end{minipage}
    \hfill{}
    \begin{minipage}[r]{1.0\columnwidth}
        \centering
        \includegraphics[width=0.75\columnwidth]{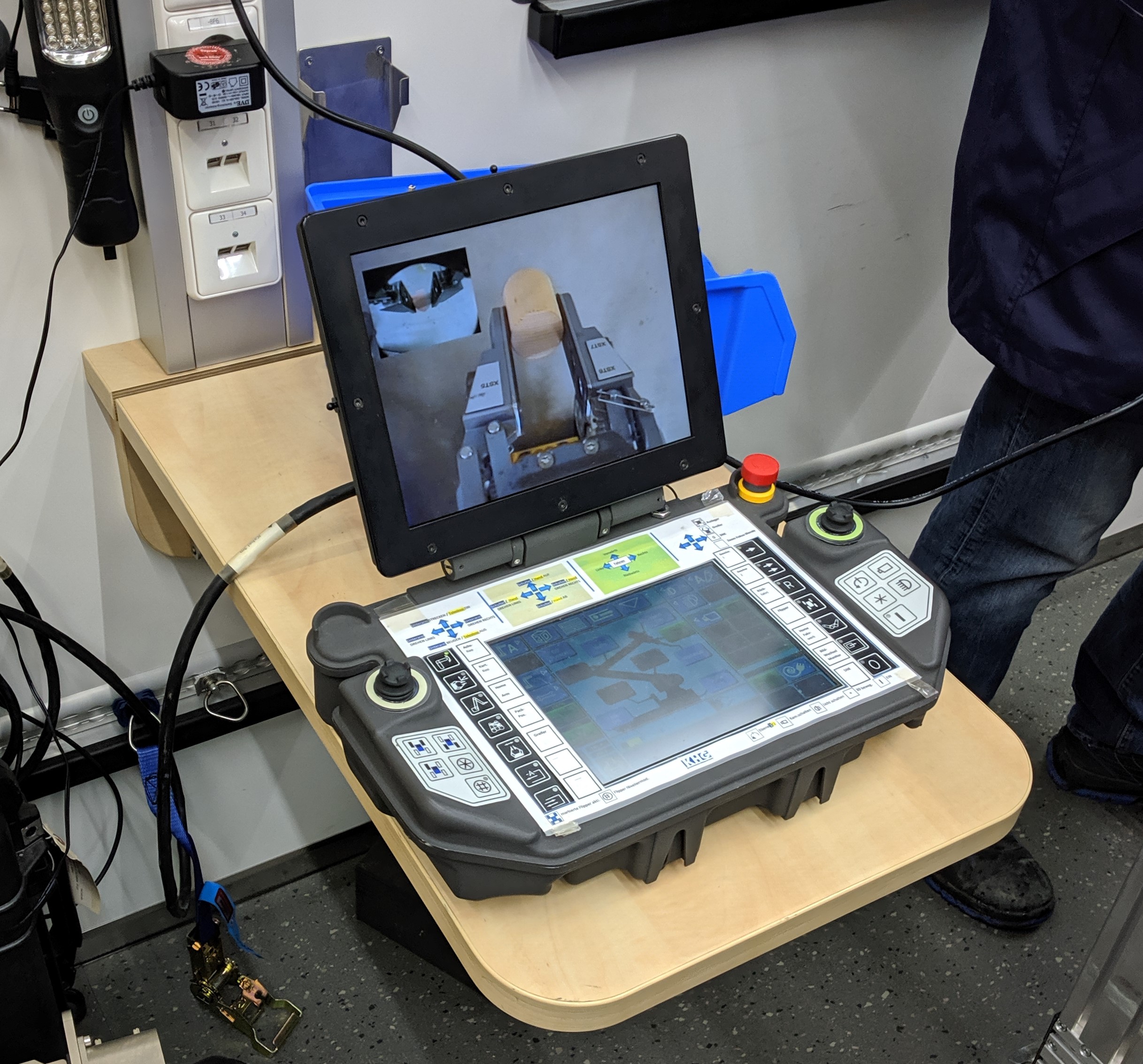}
        \caption{The Telemax OCU with a touch panel in the middle and a monitor for video feedback. It can also be connected to an external screen.}\label{fig:Telemax_ocu}
    \end{minipage}
\end{figure*}

\subsection{Robotic Apparatus}

KHG features a fleet with a wide range of robots, both Unmanned Ground Vehicles (UGVs) and Unmanned Aerial Vehicles (UAVs), covering the majority of tasks in industrial and disaster environments. The fleet consists of small and agile robots but also heavyweight vehicles. All robots are fully teleoperated by a human from a remote location and are wireless and/or tethered (i.e., cable powered). The majority of them are radiation-shielded. The camera inspection robots are particularly important for inspecting areas and pipelines. 

Complementary to the robots, a relay communication network can be deployed inside and outside a building to increase the distance at which the control center can control teleoperated robots. Inside the buildings, a relay station is deployed mounted on a tracked heavy-duty robot. For outdoor operations, a 34-meter tall telescopic mast mounted on a truck is used to support the radio link.

The reader is referred to the KHG's website\footnote{\url{https://khgmbh.de/}} for further details regarding the apparatus. Here we will further describe the equipment most relevant to the specific exercise.

\subsubsection*{Telemax (Telerob)} 
The main robot used during the field exercise was the Telemax robot (see Fig. \ref{fig:Telemax}). This robot has both retractable flippers and all-road wheels to cope with various surfaces. The chassis and the manipulator are equipped with 4 on-board cameras to cover a variety of viewing angles and offer the operator a better overall perception experience. Telemax also features a manipulator with 6 Degrees of Freedom (DoF) and an 360-degrees endless-rotation gripper. Telemax features an on-board set of 5 different sampling probes (see Fig. \ref{fig:Telemax}) designed to be easily grasped by the gripper. They are interchangeable and apply to various sampling needs. Operators can switch between probes for sampling dust particles, radioactive particles, and liquid. This is a critical functionality as these samples can be potentially crucial for the progress of a mission.

Telemax is controlled via the designated Operator Control Unit (OCU) (see Fig. \ref{fig:Telemax_ocu}) featuring a touchscreen for alternating between different modes of operation and a 10-inch display for video transmissions from the robot's on-board cameras. The modes of operation include different driving modes such as wheels only, flippers only, or both. Additional modes include the all on-board camera mode and the manipulation mode.



\subsubsection*{Mobile Control Room}
\label{mcr}
\begin{figure*}
    \begin{minipage}[l]{1.0\columnwidth}
        \centering
        \includegraphics[width=0.85\columnwidth]{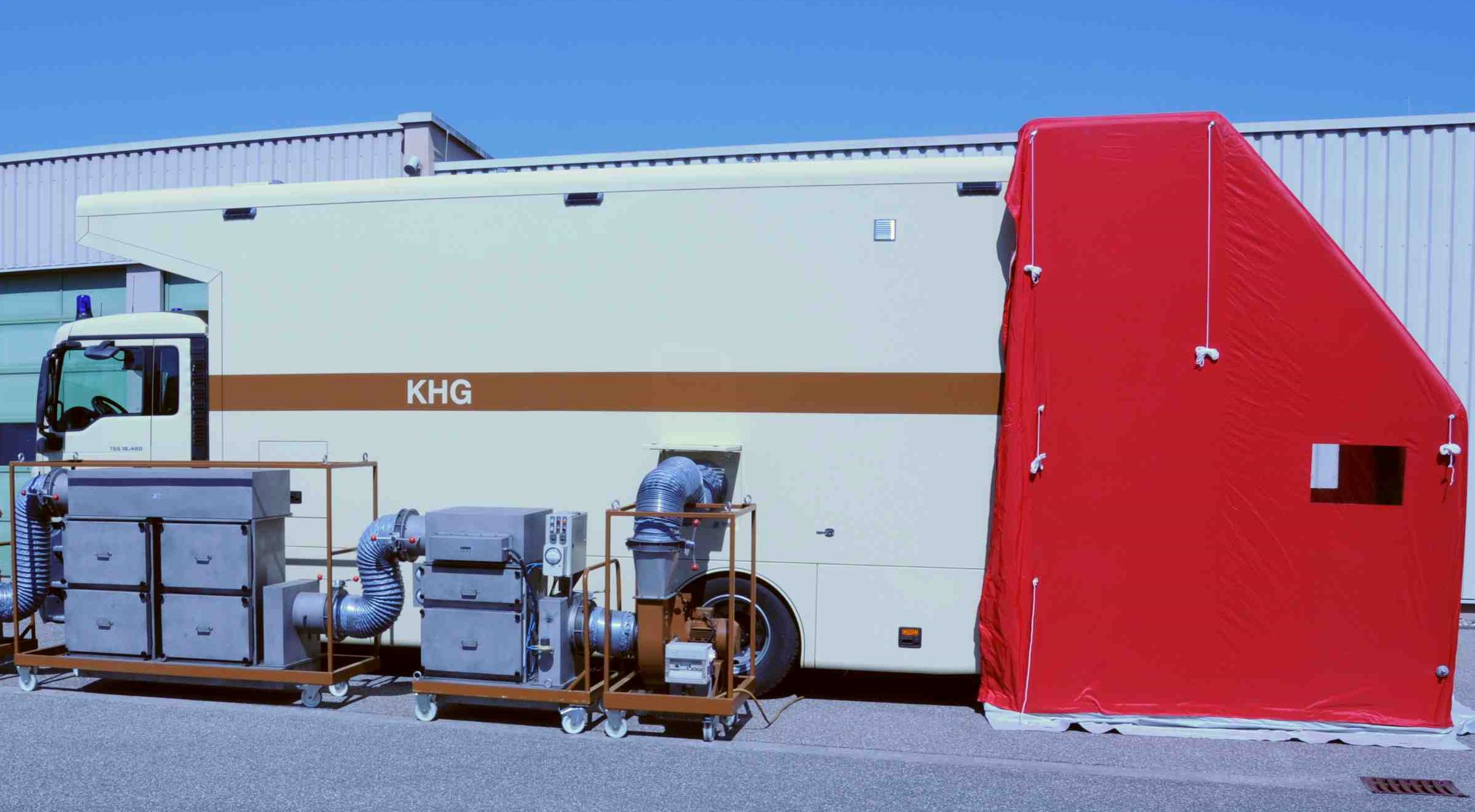}
        \caption{The truck that houses the mobile control room, including the mobile air filtering system and the red inflatable tent used as an entrance.}\label{fig:control_room}
    \end{minipage}
    \hfill{}
    \begin{minipage}[r]{1.0\columnwidth}
        \centering
        \includegraphics[width=0.85\columnwidth]{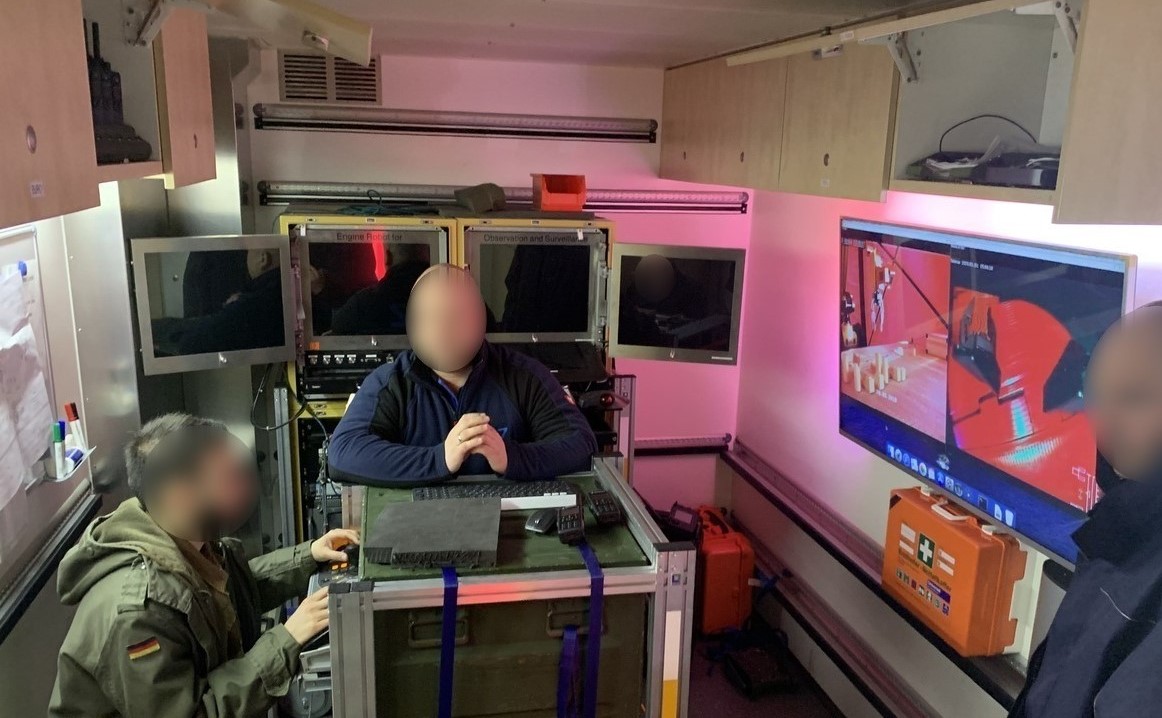}
        \caption{The various OCUs inside the mobile control room.}\label{fig:control_room_ocu}
    \end{minipage}
\end{figure*}

During robot-assisted tasks, operators are isolated in a specially designed control room on a truck, featuring various OCUs for the different robots (see Fig. \ref{fig:control_room} and \ref{fig:control_room_ocu}). The truck is designed to be strategically located during a field operation, at a safe distance for humans and equipment, and to ensure stable wireless communications. An adaptable mobile filter system that is carried along (trailer) ensures that the operators get air 99.99\% free of contamination in the control room. Access to the control room is given via the inflatable tent and the transport room, which contains an air lock function by means of 3 pressure zones.

\section{Field Exercise}

KHG has invited us as robotic researchers to participate in the training program and the exercise as robot operators. For thorough documentation of the exercise, data in the form of informal chats with the KHG personnel, photos, videos, and notes were collected before and after each mission in a systematic manner. A more formal data collection method was not followed to avoid interfering with the exercise.

The training and field exercise consisted of two main parts. At first, operators were familiarized with the equipment, including the OCUs (or in some cases the entire control room) and the robots. The second part was two realistic hands-on missions.

\begin{figure*}
    \begin{minipage}[l]{1.0\columnwidth}
        \centering
        \includegraphics[width=0.75\columnwidth]{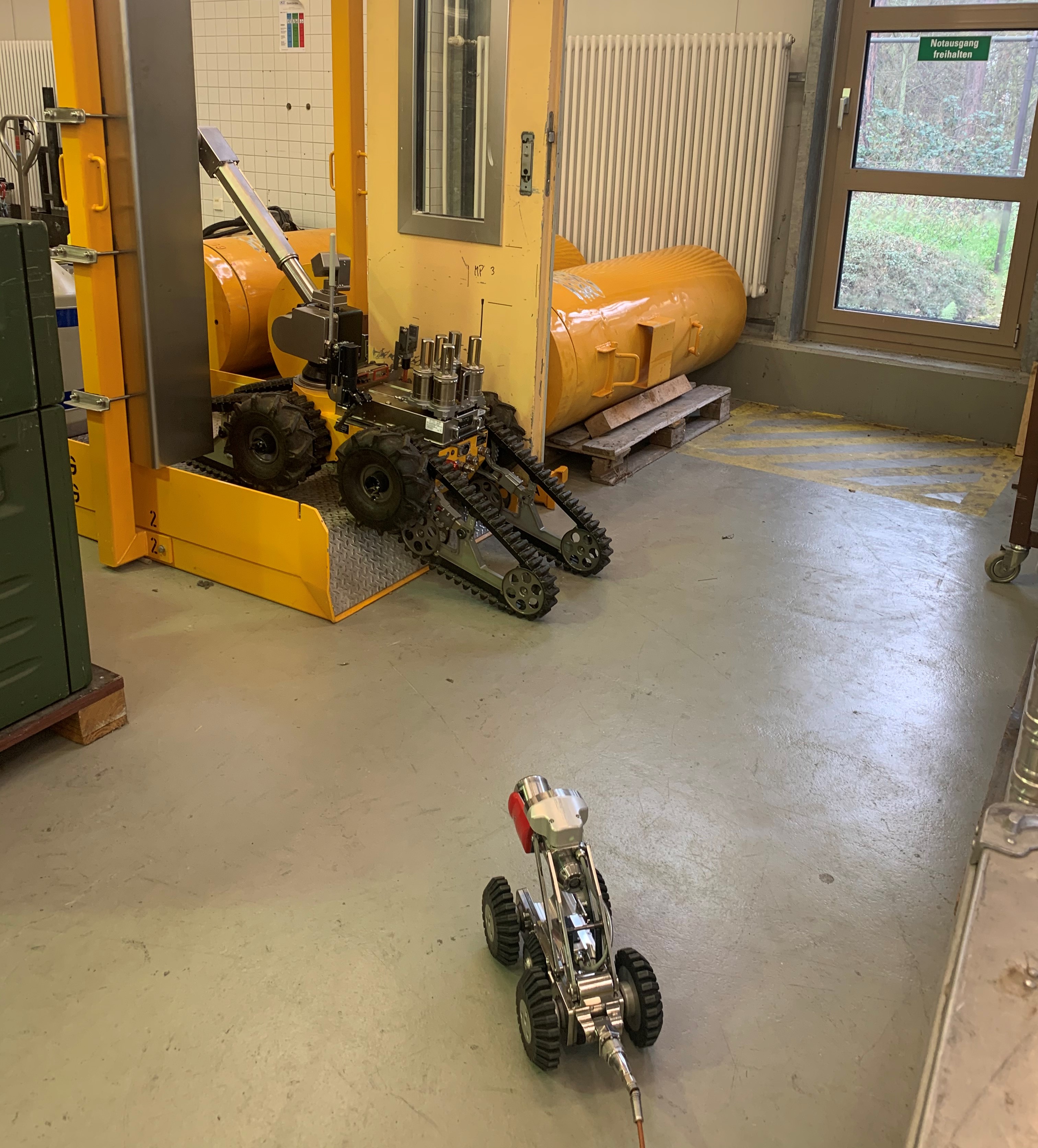}
        \caption{Telemax and the reconnaissance robot during the second mission. The second robot provides an additional viewpoint for the operator.}\label{fig:mockup}
    \end{minipage}
    \hfill{}
    \begin{minipage}[r]{1.0\columnwidth}
        \centering
        \includegraphics[width=0.85\columnwidth]{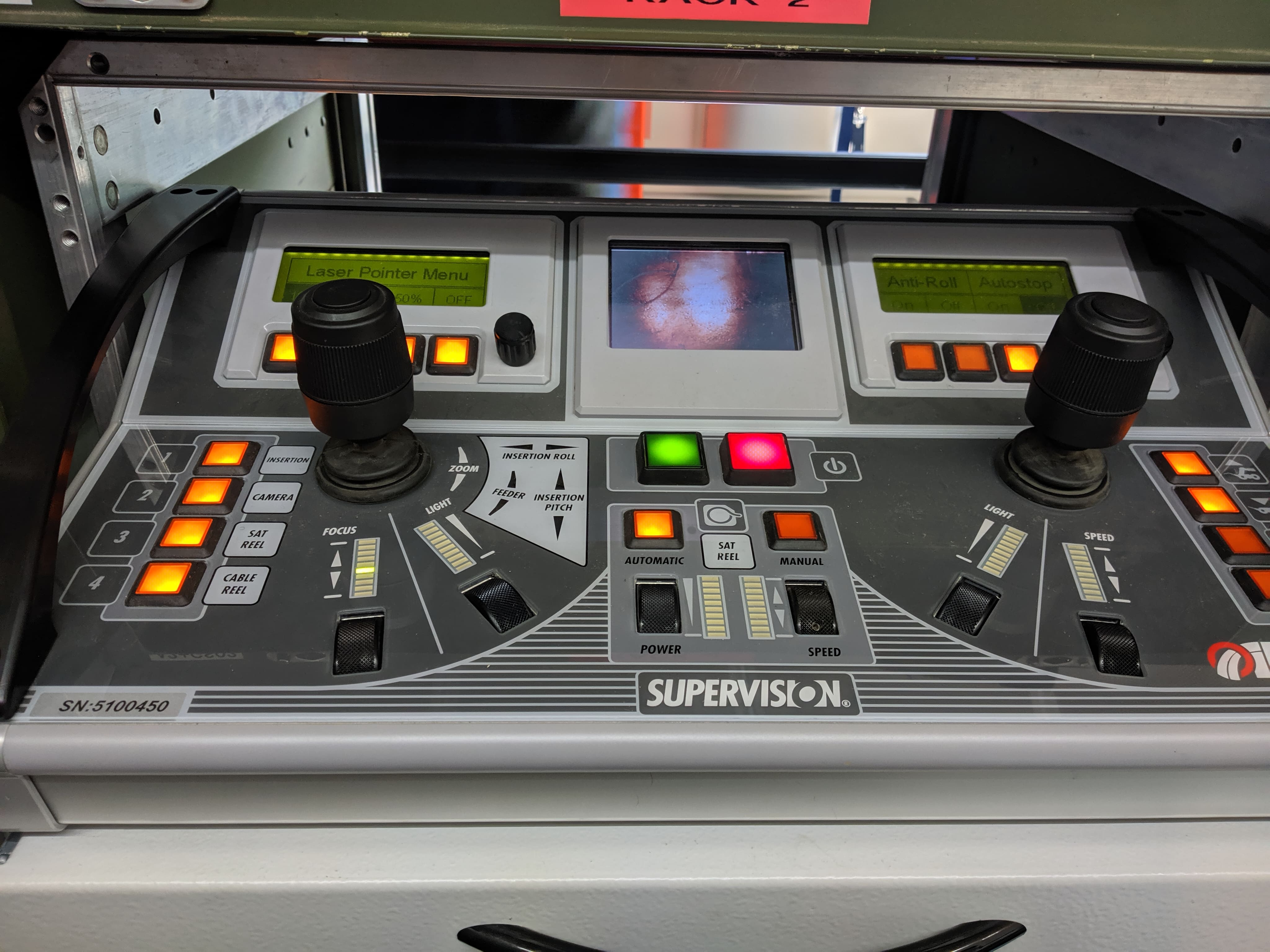}
        \caption{Reconnaissance robot's control unit. The small screen in the middle is for video feedback. It can also be connected to a larger monitor.}\label{fig:inspocu}
    \end{minipage}
\end{figure*}

The main robot for the exercise was Telemax (see Fig. \ref{fig:Telemax}), which was controlled via its OCU. Additionally, a small and agile tethered camera reconnaissance robot (see Fig. \ref{fig:mockup} and \ref{fig:inspocu}) always followed the main robot offering an external point of view and assisting the operator in gaining Situational Awareness (SA). This principle is followed by KHG in every remote robot-assisted mission and is a common practice in robot field deployments \cite{Dufek2021_best_viewpoints}. Lastly, each robot was controlled by one operator.

At first, the operators were introduced to the basic physical characteristics, the capabilities of the mobile robots, and their driving behavior. This was achieved by driving the robots in a wide-open area. Lastly, operators were familiarized with the perception during remote manipulation and the basic functions of each OCU.

\subsection{First Mission: Warm-up Navigation and Manipulation}

The first mission can be considered a preparation (i.e., \say{warm-up}) phase for the operators. The operators can get physically and mentally ready for the actual missions. Our personal experience proved this tactic helpful since better coordination between the mind, eyes, and hands can be achieved.

The task was carefully navigating Telemax inside a warehouse full of other vehicles and machinery. The initial goal was to successfully drive the robot out of the warehouse's door. Then, the operators had to drive the robot in the outer area and through a small opening in a gazebo, where a challenging manipulation task would occur. The lighting inside the gazebo was poor, so we had to use the robot's on-board auxiliary lighting devices. The goal was to collect several (6 in total) small LEGO-like wooden pieces and stack them in a way that forms a small slide. Then we had to pick a tiny steel globe and put it on the top of that slide, releasing the gripper's fingers to let the globe run down the slide. The task was difficult and stressful due to the high workload, fatigue, and multitasking. It was a combination of performance degrading factors such as poor lighting, narrow field-of-view, and lagging camera feedback. Trying to gain SA via video feedback to achieve good coordination between eyes and hands is a challenging task and requires tremendous concentration. Lastly, throughout this demanding procedure, the robot's OCU felt cumbersome to use.



\subsection{Second Mission: Narrow Space Navigation and Hazardous Materials Handling}

The second mission addressed a realistic robot-assisted nuclear emergency response scenario. The mission's objectives were to a) navigate Telemax through a hazardous area;  b) use the manipulator arm mounted on the robot to remove a contaminated object. The mission symbolized the need for safe navigation through a hazardous area where no humans could approach  and handle contaminated materials. High radiation levels and unstable environmental conditions increase the risk levels drastically, necessitating the use of robots, which can be considered \say{consumables} in such circumstances \cite{CasperMurphy2003_911}. 

The operator had to drive Telemax towards a way-point reaching a mock-up structure with a closed door (see Fig. \ref{fig:mockup}). This structure symbolizes a potential room inside a nuclear power plant where a radioactive contamination accident happened. Robot-assisted operations could prevent extensive damage and contain the contamination quickly and safely. Throughout the mission, Telemax was followed by the inspection robot to offer external video feedback according to the protocol. 

Once the Telemax reached the target door, the operator had to open the door via the door handle using a steel wire loop attached to the side of the manipulator arm. The task was demanding because once the wire loop caught the door handle, the operator had to simultaneously carefully control the manipulator and move in reverse the whole robot to leave adequate space for the door to open fully. 

\begin{figure}
    \centering
    \includegraphics[width=0.85\columnwidth]{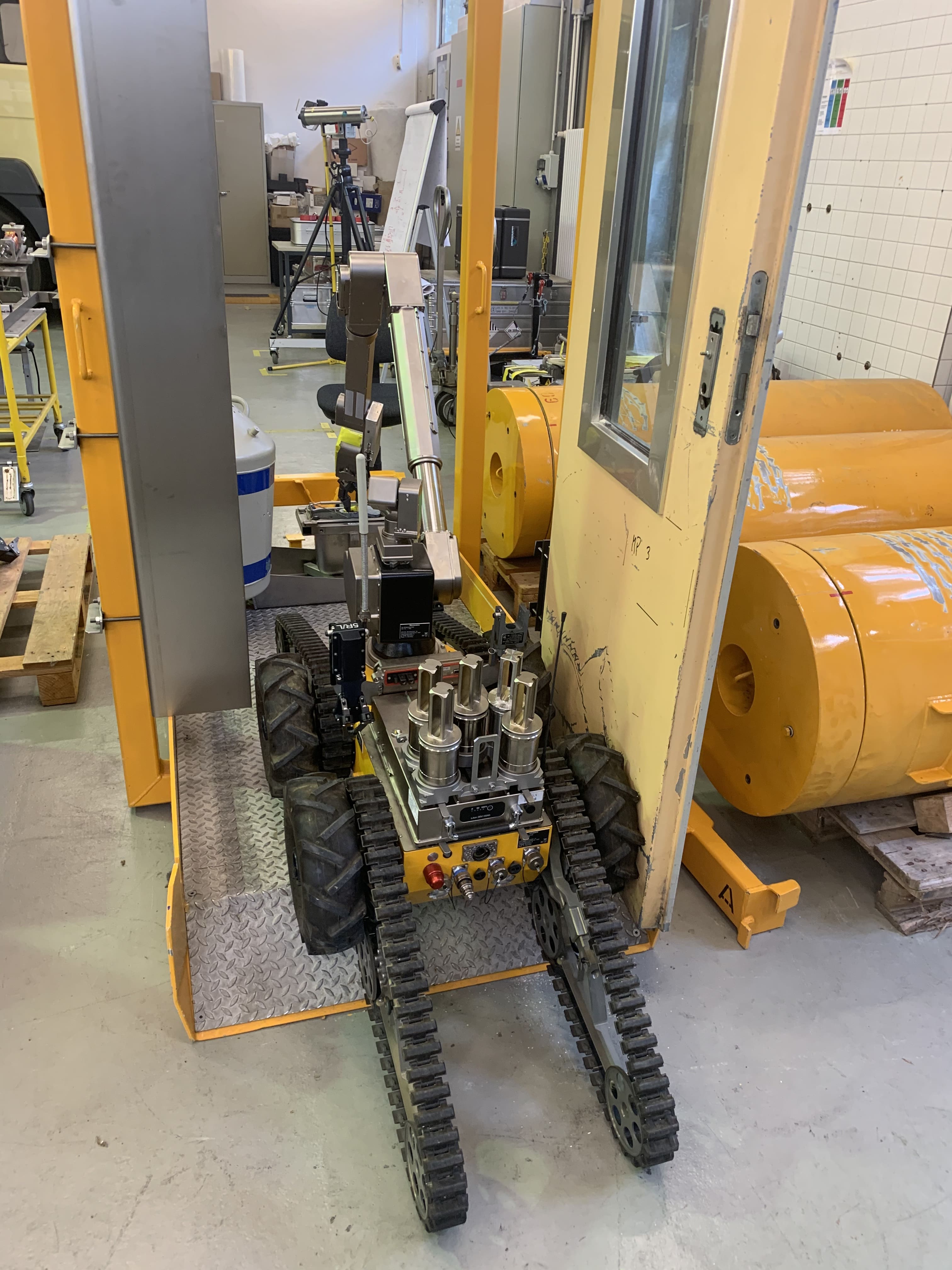}
    \caption{Telemax during the second mission while entering the mock-up structure.}
    \label{fig:mockup2}
\end{figure}

The next step was to extend the robot's flippers to go through the door, as there was a steep ramp right at the door sill (see Fig. \ref{fig:mockup} and \ref{fig:mockup2}). When the robot had safely entered the confined space where the manipulation task would take place, the operator had to retract the flippers again. 

Then, the operator had to examine the area via the multiple Telemax's on-board cameras. The goal was to spot a small cylindrical object on the floor (AA battery shaped and sized) that was hypothetically radioactive. With the use of the manipulator arm and gripper, the lid of a small shielded container (placed in the area beforehand) had to be removed. Next, the operator picked up the small hazardous object from the floor and placed it inside the shielded container. 

The mission ended with the exact procedures, but in reverse, i.e., close the container's lid, extend the flippers, back off the robot, close the door, retract the flippers, and leave the area.

Three different operators executed this mission three times, and the average execution time was about 65 minutes. The complexity of the task played a catalytic role in the speed of execution and the outcome of the mission. It is worth mentioning that 2 out of 3 trainee operators required additional assistance during task execution. One of the team members  in the task's area, acting as a spotter for safety purposes, had to instruct and guide the operator to safely remove the robot from the scene. Additionally, under the high stress and workload of the task, operators tended to make a lot of unnecessary manipulation and navigation movements. This was indicative of the cognitive and physical fatigue that increased drastically over time. The workload was high, making error avoidance difficult and requiring 100\% concentration. Despite being an exercise, it was an exceptionally stressful and intense experience.

\section{Insights and Lessons Learned}

In this section, insights and lessons learned from our participation in the exercise and our interaction with the KHG are presented:

\textbf{1. The main priority is the robustness and reliability of the platforms (both hardware and software).} From our experience during the exercise, it is clear that for KHG and other relevant stakeholders and end-users, it is essential to have robotic platforms that can run reliably upon request and non-stop around the clock. Naturally, the nuclear industry has extremely high safety standards; thus, robotic platforms must abide by these standards and specifications. Related work supporting this insight: \cite{delmerico2019current, Carrillo-Zapata2020_needs, Murphy2014_disasterBook}.

\textbf{2. First responders and related stakeholders are very cautious about adopting newly introduced technologies without any prior systematic and multifaceted evaluation.} In the case of robotics, technology that falls into this category includes different AI and autonomous capabilities along with advanced Human-Robot-Interaction (HRI) and methods of interfacing. It is worth noting that the robotics equipment that KHG uses has no AI or autonomous capabilities. Related work supporting this insight: \cite{delmerico2019current, Kruijff2012_Miradona_earthquake, CasperMurphy2003_911}.

\textbf{3. The KHG, and consequently the stakeholders, show a clear preference for modular, simple, easy-to-use, and easy-to-learn robotic systems.} This is in terms of hardware and especially in terms of interfaces, software, and controllers. Simplicity but functionality and usability are the keywords for designers and developers, with fewer, if none, unnecessary add-ons. Related work supporting this insight: \cite{delmerico2019current,Carrillo-Zapata2020_needs, CasperMurphy2003_911} 

\textbf{4. When adequately understood in-depth, real-world scenarios are often much more technically complex and intellectually challenging than typical academic lab-based scenarios.} This complexity can stimulate academic progress. Room for improvement is being born through direct contact with robotic systems in real-world scenarios. A characteristic example of how lab-based scenarios can unexpectedly differ from the field can be found in \cite{Quenzel2021_MBZIRC_challenge}.

\textbf{5. The main impediment in performance comes from the cognitive and attentional demands on the operator.} Based on our experience in the exercise, we identified two sources: perceptual demands and task demands. This is in accordance with Murphy's proposed cognitive and attention resource demands model in field robotics \cite{Murphy2019_off_normal}. In this context, perceptual demands stem from the plain cameras' feedback with narrow fields of view and fluctuations in the quality of image transmission and lighting, resulting in reduced perception. Task demands stem from the difficulty of manually controlling the robot and especially the high DoF manipulator to accomplish the exercise's challenging tasks. Perceptual and task demands act in synergy and concurrently as the operator must operate the robot and gain SA simultaneously. This synergy contributes significantly to how quickly the operator shows signs of high workload and fatigue, leading to performance degradation. This insight agrees with a plethora of similar findings in related work \cite{Kruijff2012_Miradona_earthquake, Murphy2005_Up_from_the_rubble,CasperMurphy2003_911, fuku_blog, Chen2007_teleop_issues_interfaces}.

\section{Suggestions for Advancing Disaster Response Robotic Systems}

\subsection{Adopting the Variable Autonomy Paradigm}
In disaster response and remote inspection, robots can significantly benefit from AI and autonomous capabilities. As the current impediment in performance comes from the intrinsic difficulties of teleoperation, i.e., the cognitive workload of the operator, robots need to be capable of actively assisting the operator with task execution. The reported insights and suggestions in this paper contribute further to the field studies that have already pointed out the need for robots with autonomous capabilities that actively assist operators \cite{CasperMurphy2003_911, Murphy2005_Up_from_the_rubble, Kruijff2012_Miradona_earthquake,Norton2017_DARPA}.

Towards this end, there is an increasing amount of research aimed at advancing the AI capabilities of robots in remote inspection and disaster response tasks \cite{delmerico2019current}. Frequently this research is focused on tackling the perceptual demands of robot operation via improving the amount and quality of information given to the operators. Improvements to the OCU, SLAM, and 3D mapping methods are commonly considered \cite{delmerico2019current, Kruijff2014_NIFTi, Kruijff-Korbayova2015_TRADR}.

Enhancing the operators' perception is crucial as it can help reduce the mental effort required to acquire SA. However, it does not directly address the task demands identified here as a major performance issue. Task demands in this context mean the burden of manually controlling the robot and the errors it brings if an operator is overloaded, something enhanced perception does not directly tackle.  

We argue that research should address the manual control (i.e., pure teleoperation) problem by careful integration of autonomous control capabilities via a Variable Autonomy (VA) paradigm (also known as adjustable, sliding, or shared autonomy). In a VA paradigm, control of the robot can be switched dynamically between the robot's AI (or an AI agent) and the human in the continuous or discrete spectrum of full teleoperation to full autonomy \cite{Chiou2016_IROS_HI}. 

Specific autonomous capabilities or Levels of Autonomy (LoA), such as semi-autonomous navigation \cite{Bellicoso2018_legged_anymal} or autonomous camera control, \cite{Valiton2021} can be used on-demand to assist an operator who may be struggling to cope with issues such as high workload, multitasking, intermittent communications, and fatigue. For example, shared control assisted navigation \cite{Pappas2020_VFH_shared_control} can be activated to aid an operator in driving the robot safely through hazardous areas. Similarly, other circumstances might necessitate activating pure teleoperation for some fine-grained movements or the human reacting to unforeseen failures \cite{Murphy2008_shared_role_model}, or while in a safe area, autonomous navigation can be used to allow the operator to multitask (e.g., gaining SA) \cite{Chiou2021_THRI}. Manipulation and grasping is another task that the operator can significantly benefit from AI control assistance \cite{adjigble2019assisted} as it was one of the most challenging parts of the exercise.

Besides its practical advantages of improving performance  \cite{Chiou2021_THRI, Chiou2016_IROS_HI} and adding resilience \cite{Murphy2008_shared_role_model}, VA offers additional benefits. Researchers often assume that a full autonomy paradigm is what is required. Despite its advantages, the full autonomy approach suffers from some significant limitations \cite{Murphy2008_shared_role_model}. It assumes that the robot's AI is fully capable of performing the task with minimum or no human intervention. However, disaster response usually occurs in unpredictable and highly unstructured environments where unforeseen circumstances might necessitate uniquely human abilities (for the foreseeable future) such as critical decision making or communication with victims in a search and rescue. Additionally, there are legal and ethical concerns about adopting full autonomy in such applications. The above reasons highly contribute to the prevalent no or minimum autonomy with a human-in-the-loop paradigm \cite{Murphy2004_rescue_HRI_workflow} and towards stakeholders being conservative and not trusting the adoption of autonomous systems. VA can offer the careful and gradual integration of autonomy with the human-in-the-loop that stakeholders are looking for \cite{delmerico2019current, Murphy2009_mines, Carrillo-Zapata2020_needs}; as  professional robot operators at KHG noted in our chat during the exercise. It provides the middle ground between pure teleoperation and full autonomy \cite{Murphy2008_shared_role_model} by using the complementing capabilities of AI agents and humans \cite{Music2017}.

Lastly, from an ethical and legal perspective, Methnani et al. \cite{Methnani2021_Let_me_take_over} argue that VA ensures meaningful human control by satisfying the core values in ethical guidelines: accountability, responsibility, and transparency. This is because VA needs to answer which aspects of autonomy are adjusted by whom, how, why, and  when.

\subsection{Two-way Transfer of Knowledge Between Researchers and Stakeholders}

The two-way transfer of knowledge should aim to develop robotic capabilities that are a) relevant and useful to the end-users; b) robust and reliable after systematic evaluations in the field; and c) rely on realistic assumptions. Towards this end, we argue that researchers should aim to create or take opportunities to directly interact with stakeholders in the related field environment, e.g., taking part in realistic exercises. Our experience participating in this exercise was valuable for getting a picture of the needs, requirements, and constraints. We encourage researchers and stakeholders to engage in more such interactions. The establishment of the German Rescue Robotics Center (DRZ) \cite{Kruijff-Korbayova2021_DRZ_center} is an example of moving towards a two-way transfer of knowledge in robot-assisted disaster response. It is of great importance to facilitate and encourage the adoption and adaptation of research in applied practices in the field.  For these purposes, stakeholders can facilitate the evaluation of both existing and novel robotics capabilities in a safe but realistic manner, e.g., by allowing the participation of researchers and their systems in some of the exercises. Such practice will tackle the lack of trust that often leads to robots being deployed solely via pure teleoperation despite them having some autonomous capabilities \cite{Nagatani2013_Fuku_QuinceRobot, DeGreeff2018_industrial_exercise}. 
In addition, the active participation of FRs in research projects can make them familiar with robotic technologies such as in \cite{Kruijff-Korbayova2015_TRADR, Kruijff2014_NIFTi}. This increases the likelihood that these FRs will use the AI capabilities of their robots, such as semi-autonomous control and/or 3d mapping in the field, as shown in \cite{Kruijff-Korbayova2016_Amatrice_earthquake, Kruijff2012_Miradona_earthquake, Surmann2021_fireBerlin}. 

Lastly, the increasing number of robotic competitions is a positive step towards bridging the gap between researchers and stakeholder requirements. As suggested by Schneider and Wildermuth \cite{Schneider2019}, various competitions can be a good way to motivate researchers to develop systems that could cope with the requirements of disaster response deployments. Many competitions mainly focus on autonomy and thus allow very limited or no interaction with a human operator. We suggest that emphasis should also be given to HRI, Human-Robot Teaming (HRT), and targeted use of VA, i.e., using a combination of teleoperation and autonomy to maximize task performance and overcome unforeseen situations. This can be achieved by a) updating the competition rules to explicitly reward (e.g., higher score) teams that use systematic approaches to VA, HRI, and HRT; b) benchmark vital mission functionalities with humans-in-the-loop; and c) establishing specific tasks on HRI and HRT that the teams must compete on.



\section{Conclusion}

This paper reported on a robot-assisted nuclear disaster response field exercise. The authors were invited to observe and participate in the exercise both in their capacity as researchers and as robot operator trainees. The main contributions of this paper are the insights and suggestions aiming at the two-way transfer of knowledge between stakeholders and academia. This bidirectional flow of knowledge and experience can benefit both sides extensively via the targeted development of new robotic technologies that end-users can adopt and deploy.

Based on our experience, the main impediment in performance comes from the cognitive and attentional demands on the operator. Furthermore, robotic platforms and modules should aim to be robust and reliable in addition to their ease of use and training. This can make the stakeholders more open to testing or even adopting novel systems. Focusing on more realistic conditions when designing and testing new systems is a good starting point toward implementations that are appealing to the stakeholders. 

Last, our primary suggestion for improving autonomous capabilities in disaster response is adopting a VA paradigm. It could offer a gradual integration of autonomy with the human-in-the-loop as it provides the middle ground between teleoperation and autonomy, leading to a potential increased end-user acceptance. Additionally, as VA can directly alleviate some of the operator's control burden, it has the potential to reduce the operator's workload and offer flexibility with the on-demand use of autonomous capabilities. Thus, improving the robotic system efficiency and overall mission success. 

Ultimately, the adoption of robotic technologies in field deployment is determined by the real-world requirements and needs of stakeholders/end-users.   


 \section*{Acknowledgment}
 The authors would like to thank KHG for their hospitality, for inviting us to participate in the exercise, and for permitting us to publish this material.

\bibliographystyle{IEEEtran}
\bibliography{IEEEabrv, refs.bib}

\end{document}